\documentclass{article}
\pdfoutput=1

    \usepackage[preprint]{neurips_2024}

\usepackage[utf8]{inputenc} %
\usepackage[T1]{fontenc}    %
\usepackage{hyperref}       %
\usepackage{url}            %
\usepackage{booktabs}       %
\usepackage{amsfonts}       %
\usepackage{nicefrac}       %
\usepackage{microtype}      %
\usepackage{xcolor}         %

\usepackage[textsize=tiny]{todonotes}

\newcommand\mvp{\vspace*{-0.1in}}
\usepackage{listings}
\usepackage[most]{tcolorbox}

\def\mund#1{\smallskip\noindent{\bf #1: }}
\newcommand{\mytcbinput}[4]{
\tcbinputlisting{
      listing file=#1,
      breakable,
      title=#2,
      label=lst:#4,
      listing only
    }
}
\usepackage{array}

\newcolumntype{P}[1]{>{\centering\arraybackslash}m{#1}}
\usepackage{multirow}

\title{On the Self-Verification Limitations of Large Language Models on Reasoning and Planning Tasks}

\author{%
  Kaya Stechly\\
  Arizona State University\\
  \texttt{khstechl@asu.edu} \\
  \And Karthik Valmeekam\\
  Arizona State University\\
  \texttt{kvalmeekam@asu.edu} \\
  \And Subbarao Kambhampati\\
  Arizona State University\\
  \texttt{rao@asu.edu} \\
}

\begin{document}

\maketitle
\begin{abstract}
There has been considerable divergence of opinion on the reasoning abilities of Large Language Models (LLMs).
While the initial optimism that reasoning might emerge automatically with scale has been tempered thanks to a slew of counterexamples--ranging from multiplication to simple planning--there persists a wide spread belief that LLMs can self-critique and improve their own solutions in an iterative fashion.
This belief seemingly rests on the assumption that verification of correctness should be easier than generation--a rather classical argument from computational complexity--which should be irrelevant to LLMs to the extent that what they are doing is approximate retrieval.
In this paper, we set out to systematically investigate the effectiveness of iterative prompting in the context of reasoning and planning.
We present a principled empirical study of the performance of GPT-4 in three domains: Game of 24, Graph Coloring, and STRIPS planning.
We experiment both with the model critiquing its own answers and with an external correct reasoner verifying proposed solutions.
In each case, we analyze whether the content of criticisms actually affects bottom line performance, and whether we can ablate elements of the augmented system without losing performance. We observe significant performance collapse
with self-critique and significant performance gains with sound external verification.
We also note that merely re-prompting with a sound verifier maintains most of the benefits of more involved setups. 
\end{abstract}

\section{Introduction}

Large Language Models (LLMs) have rapidly captured the attention of the AI research community with their exceptional natural language completion capabilities.
Trained on web-scale language corpora, these models have demonstrated the ability to generate seemingly valuable completions across a wide range of topics.
This has led to a surge of interest in determining whether such models are able to perform well on reasoning tasks.
Though initial anecdotal results showed promise, further systematic studies revealed inconsistencies and significant issues when applied to reasoning tasks--such as simple arithmetic or logic \cite{dziri2023faith} and planning \cite{valmeekam2023on}. These results questioned the robustness of their reasoning abilities and led researchers to explore ways to augment and improve these systems.

Of particular interest to us are emerging claims about LLM self-critiquing. In a typical setup, an LLM is iteratively prompted so that it both generates candidate solutions and, in response to separate queries, evaluates them. This process is looped until some stopping condition in hopes of potentially generating a refined answer. 
Current works \cite{yao2023tree, shinn2023reflexion, weng2023large, chen2023teaching, madaan2023self}, while admitting that LLMs are not good reasoners on their own, still exhibit considerable optimism about such self-critique systems. This belief seems to rest largely on the assumption that verification of correctness should be easier than generation for many reasoning problems--a rather classical argument from computational complexity. However, we think there are grounds to be skeptical of this assumption. The complexity of the reasoning task should 
be largely irrelevant to LLM performance, especially if what they are doing is better modeled as approximate retrieval \cite{kambhampati2024can}. 

Intrigued by the prevailing optimism, in this paper we set out to systematically investigate the effectiveness of using LLMs to critique their own generations in the context of planning and reasoning. To gain meaningful insights into the verification/critiquing abilities of LLMs for reasoning tasks, it is crucial to test them on \textit{formal} tasks--ones where machine-verifiable ground truths are available for both generation and criticism. Note that such verification is not feasible in style-based/qualitative tasks like creative writing \cite{yao2023tree} where there is no formal notion of correctness and the critique can vary widely. With this understanding, we select three distinct reasoning problems: \textit{Game of 24}, \textit{Graph Coloring}, and \textit{STRIPS planning}, in which there exist formal notions of correctness that allow us to automatically check the veracity and quality of both (binary) verification and critique generated by the LLM. 

Our methodology employs a system (which we refer to as LLM+LLM) that uses the same LLM (the state-of-the-art GPT-4 \cite{achiam2023gpt}) for iterative solution and verification/critique generation.
A generation prompt is sent to the LLM. Its response is used to create a verification prompt, which is then sent back to the same LLM.
We use the feedback generated in this way to then create a backprompt, thus restarting the cycle.

Across almost all of our domains, this self-verification system {\em worsens} performance. We find that as the number of backprompts increases, this kind of self-correction consistently degrades output quality. Our analysis reveals that the verifier LLM's false negative rate is significant across our domains. In essence, even when the LLM generates a valid solution, the verifier LLM rejects it often enough that overall performance suffers.

We contrast this performance collapse with two baselines. The first is an ablated variant of the system (which we refer to as LLM+Sound Verifier), where an external sound verifier evaluates the LLM's generations and produces critique.
This setup gives substantial performance gains across all domains, but closer analysis shows that the level of feedback often doesn't seem to matter--as long as the verifier is sound, improvement remains regardless of how much or how little feedback the LLM receives.

We ablate the system further, and remove critique entirely. In this setup, the LLM is repeatedly queried with the exact same base prompt until a sound verifier certifies its solution.
Within this impoverished setting, prompts no longer maintain a past history of guesses, yet we can maintain most, if not all, of the gains shown by our previous, more complicated, more expensive setups.

Our empirical results suggest that the benefits of iterative prompting and verification can easily be misattributed to opaque self-critique and seemingly rich feedback.
Thus, future implementations of LLMs for reasoning tasks should take the form of LLM-Modulo systems \cite{kambhampatirole} where verification is done by external sound systems.
In the rest of the paper, we first review related work and discuss domain backgrounds. Then, we explain our methodology, and finally closely analyze LLM self-verification abilities on our domains.

\mvp 
\section{Related Work}\label{sec:related}
\mvp 

Following the release of GPT-4, anecdotal accounts of performance on reasoning tasks  \cite{bubeck2023sparks} spurred much research into the capabilities of LLMs on various reasoning domains, from a menagerie of benchmarks covering basic problems \cite{kojima2022large} to planning \cite{valmeekam2023on}, logic and arithmetic \cite{dziri2023faith}, analogical reasoning \cite{webb2023emergent}, and even math puzzles \cite{yao2023tree}.
Though these seemed initially promising, systematic studies began to generate negative results across many of these domains \cite{valmeekam2023on, silver2022pddl, abdin2023kitab, ullman2023large, gendron2023large}, claiming that LLM scaling shows much lower returns for reasoning tasks \cite{rae2021scaling}, and showcasing brittle performance in the face of minor problem permutations \cite{mccoy2023embers, dziri2023faith, arakelyan2023exploring}.  

In response, researchers created augmented systems which embed the LLM in larger frameworks in an attempt to improve performance.
These take many forms: common search algorithms with the LLM cast in the role of heuristic \cite{yao2023tree, hao_reasoning_2023},
approaches which reduce error rates by enforcing various consistency conditions \cite{du2023improving, cohen2023lm, jiang2023backward},  
and direct LLM self-critique \cite{shinn2023reflexion, weng2023large, chen2023teaching, huang2022inner, madaan2023self,yao2022react}.

In the current work, we are interested in examining this third approach: LLM self-critique. In the most basic case, the LLM is queried for an answer, and then is re-queried with its own response together with some instructions to critique or improve it, with this process looped until some stopping condition.
This is fundamentally based on the intuition that verification is easier than, or at least different enough from, generation that such a process can improve performance--in analogy to human self-critique\cite{weng2023large}.

The literature abounds with strong, well-cited, and well-referenced claims about the efficacy of these techniques. \cite{shinn2023reflexion} claims there is an “emergent property of self-reflection in LLMs” and that “self-reflection is extremely useful to learn complex tasks over a handful of trials.” Their experiments claim that every variety they try leads to improvement, and that this is because “self-reflective feedback acts as a ‘semantic’ gradient signal by providing the agent with a concrete direction to improve upon, helping it learn from prior mistakes to perform better on the task.”\footnote{Note that our results ablate away much of this signal (especially the ‘concrete direction’ that exists in explicit critique) to find that most of the improvement in our domains comes from the soundness of the verifier.}
Other works claim this self-correction does not require “any human feedback” \cite{chen2023teaching} and that 
“even when an LLM cannot generate an optimal output on its first try, the LLM can often provide useful feedback and improve its own output accordingly,” \cite{madaan2023self} seeming to indicate that these claims generalize beyond the domains, problems, and prompts they were originally made for.

However, some further systematic investigations have found less impressive results in logical fallacy detection \cite{hong2023closer} and HotpotQA \cite{huang2023large}, demonstrating very brittle improvement at best, some of which could be replicated sans self critique by merely including missing domain-general information into the original prompt. The authors of the CRITIC framework\cite{gou2023critic} were the first to notice that, in some cases, LLM self-critique can lead to decreases in performance when compared to sound verification. Contemporaneous to our work,\footnote{Preliminary results from our work were originally presented in two papers at a NeurIPS 2023 workshop.} \cite{huang2023large} investigate two-round self-correction schemes in the GSM8K, CommonSenseQA, and HotpotQA domains. They compare which answers were changed (from correct to incorrect or incorrect to incorrect) and which weren't, and discuss extensions of their argument to multiagent debate.

Our own work focuses on autonomous multi-round self-verification within three formally verifiable domains that reflect reasoning tasks. We extend previous work by ablating the self-critique system thoroughly to pinpoint the source of performance deterioration, considering more prompting rounds (up to 15), and by examining a new set of domains which we argue are better and more broadly applicable tests of reasoning and self-correction capability.

Reasoning is a fraught term. Previous work has used it to refer, among others, to the  human ability to draw conclusions\cite{leighton2004nature}, to the ability to apply common sense to simple scenarios, to positive performance on short-form written tasks, and to formal deductive inference. However, it is often unclear which definition a given set of authors presupposes when making claims about LLM reasoning capabilities.
This muddies the discussion and contributes to a strange duality: highly cited papers claim that LLMs are general-purpose reasoners \cite{kojima2022large, wei2022chain, bubeck2023sparks, zhou2022least}; that they have strong, human-like self-reflection capabilities which allow them to correct reasoning mistakes they do make \cite{shinn2023reflexion, chen2023teaching, madaan2023self}; that they can answer difficult, never-before-seen questions via in-context learning as long as they are allowed to use chain of thought to generate intermediate scratch work \cite{dong2022survey}; that they can pass or come close on many high school and college-level examinations \cite{achiam2023gpt, gilson2023does, raimondi2023comparative, thaker2024large, yeadon2023exploring, de2023can, kortemeyer2023could} and that performance on such standardized exams is evidence about their reasoning capabilities and domain expert knowledge \cite{wang2023scibench}.
Yet, responses to counterexamples and negative results, anecdotally, fall back on a much weaker, seemingly contradictory constellation of premises: LLMs only perform well on things they were trained on, and--in fact--if a model performs poorly, we can only conclude it wasn't trained on that (but if it performs well, it is generalizing); the average non-expert human would fail on this task if presented it with zero context or training, therefore it's unsurprising that the LLM fails; no good prompt engineer would query the LLM in this fashion. (How to tell if a prompt is good? It follows one of several anthropomorphized design patterns and, most importantly, the result is positive.)

These shifting definitions and implicit assumptions make it very difficult to make concrete claims and expect to be understood, and they make it even more difficult to pin down claims made by others or attempt to falsify them. In the current work we address this by restricting our focus to fully specified, formally verifiable problems which can be solved directly by deductive methods. Though these may at first seem like a very narrow class, especially when compared to the cornucopia of commonsense, language-based, and domain-specific benchmarks in the literature, we argue that they are fundamental, as any other reasoning tasks must include components that test these same capabilities--otherwise they are merely testing recall.
Our work extends studies that have looked at similar problems, especially those that examined LLM planning capabilities and other classical reasoning problems \cite{valmeekam2023planbench, valmeekam2023on, stechly2024chain, dziri2023faith}. 
However, no previous work has looked carefully at a broad range of formal verification problems. Filling in this gap is important, as a lack of benchmark coverage contributes to the illusion that LLMs possess greater competency than they really do \cite{saxon_benchmarks_2024}.

Furthermore, common domains fall short for evaluating the reasoning and self-critique abilities of LLMs for additional reasons: test set memorization, lack of problem difficulty, and lack of ground truth.

\mund{Test set memorization} Due to the black box nature of state of the art models, ensuring that they weren't trained on those problems is difficult, and there is compelling evidence that they have memorized significant chunks of common benchmark sets \cite{roberts2023data}. Many benchmark sets do not allow for arbitrary generation of novel questions, or worse, draw data from publicly available sources--the same sources LLM trainers have access to \cite{yang2018hotpotqa, srivastava2023beyond}.
We consider arbitrary generation of new instances of varying difficulty a key desideratum for any evaluation domain.

\mund{Lack of problem difficulty} Some of the benchmarks (e.g. HotPotQA \cite{huang2023large}, GSM-8k \cite{madaan2023self}, %
) used in evaluations of self-verification are easy--that is, SoTA LLM performance is already high--and are therefore much less informative about the effects of the refinement procedure. Additionally, many such sets over-constrain the solution space, usually by putting the question into multiple choice format. Not only does this make valuable and interesting critique hard to produce and evaluate, but it trivializes refinement: even a very simple agent can solve an $n$-choice problem with $n-1$ critiques--just don't repeat the same answer. Conclusions drawn over reduced problem spaces of this type are unlikely to generalize.

\mund{Lack of ground truth} A number of tasks that LLMs are evaluated on (e.g. writing prompts \cite{yao2023tree}, constrained text generation \cite{lin2020commongen}, toxicity control \cite{welleck2022generating, gou2023critic}  %
, etc.) are problems without a well-defined ground truth. Instead, they are evaluated by a couple of indirect %
methods. Some require an assorted set of metrics which may not be well-validated for LLMs (e.g. see \cite{ullman2023large} for discussion on problems with transferring results from human-validated tests). Some are scored by humans \cite{yao2023tree}. And some are evaluated by another pre-trained language model \cite{madaan2023self} or black box AI evaluator \cite{welleck2022generating}. This makes conclusions much harder to draw.

\section{Background On Test Tasks}

We evaluate GPT-4's self-critique abilities over three distinct tasks, chosen because we believe they are good proxies for harder reasoning tasks, and because they allow freedom in arbitrary generation of additional instances 
while providing easy-to-deploy formal verifiability and guaranteed quality.

This gives more than just flexibility--it also decreases the chance that our instances are represented in the black box model's opaque training sets. This strengthens our results by reducing the likelihood that the model can substitute approximate retrieval for general reasoning ability. %

Any given problem in our sets also has the property that it has a large number of potential solutions, and this solution space cannot be substantially reduced through simple pattern-matching. As we are interested in self-verification loops where the LLM has access to its previous guesses, it is very important that removing a handful of possible solutions does not trivialize the problem. Compare this to common multiple choice question datasets, where any $n$-option problem can be solved in $n$ exclusive guesses. %

\subsection{Game of 24}
\textit{Game of 24} is a math puzzle where the goal is to combine four numbers with parentheses and basic arithmetical operations (addition, multiplication, subtraction, and division) to create an expression that evaluates to 24. The numbers are typically constrained to the range 1-12, a nod to game's playing card roots. Previously, it has been used as a domain of evaluation for other LLM self-verification schemes (\cite{yao2023tree} and fulfills our domain desiderata (see \ref{sec:related}). We use it here to enable direct comparisons between previous work and the current paper. 

Following \cite{yao2023tree}, we use data scraped from \url{4nums.com}. This list of problems is ordered from shortest to longest average human solution time. Like \cite{yao2023tree}, we evaluate our generation tasks on instances 901-1000. However, when evaluating verification and critique alone, we use instances 1-1000.

Verification in this domain is straightforward: given a proposed expression, simplify it using basic arithmetic and check if it is equal to 24.
As a sound verifier, we use \verb|SymPy|\footnote{https://www.sympy.org/en/index.html}, a common Python library for symbolic mathematics, and handle any errors that it throws (for instance, if there are unbalanced parentheses) by outputting feedback that says the LLM's generation was malformed.

\subsection{Graph Coloring}
\textit{Graph coloring} is a a canonical NP-complete reasoning problem that is related to both propositional satisfiability as well as practical problems like scheduling and allocation. The complexity class NP contains problems that are hard to solve, but easy to verify, so this allows our It is broad enough to give insights into reasoning more generally, yet simple enough that it can be specified and evaluated by a human or basic pattern matching.

In this work, an instance of a graph coloring problem consists of a planar graph together with an optimal coloring number $n$. The goal is to output a solution that assigns one of $n$ colors to each vertex such that no two edge-connected vertices share a color.

Using \verb|GrinPy|\footnote{https://pypi.org/project/grinpy/} to handle common graph operations, we built a test set of 100 graphs of varying small sizes. Each graph was constructed using a variant of the Erdős–Rényi method (\(p=0.4\)), with any non-planar or repeat graphs discarded. These were compiled into the standard DIMACS format \cite{dimacsrutgers} together with the graph's precalculated chromatic number.

Verifying that a proposed coloring is correct is also easy: just check the colors of every edge. If any of them has two vertices of the same color, reject the coloring.
Our sound verifier is a simple, single \verb|for|-loop implementation of this idea in Python: for each edge in the graph description, we check that both of its vertices are different. 
\begin{figure*}[ht]
    \centering
    \includegraphics[width=\textwidth]{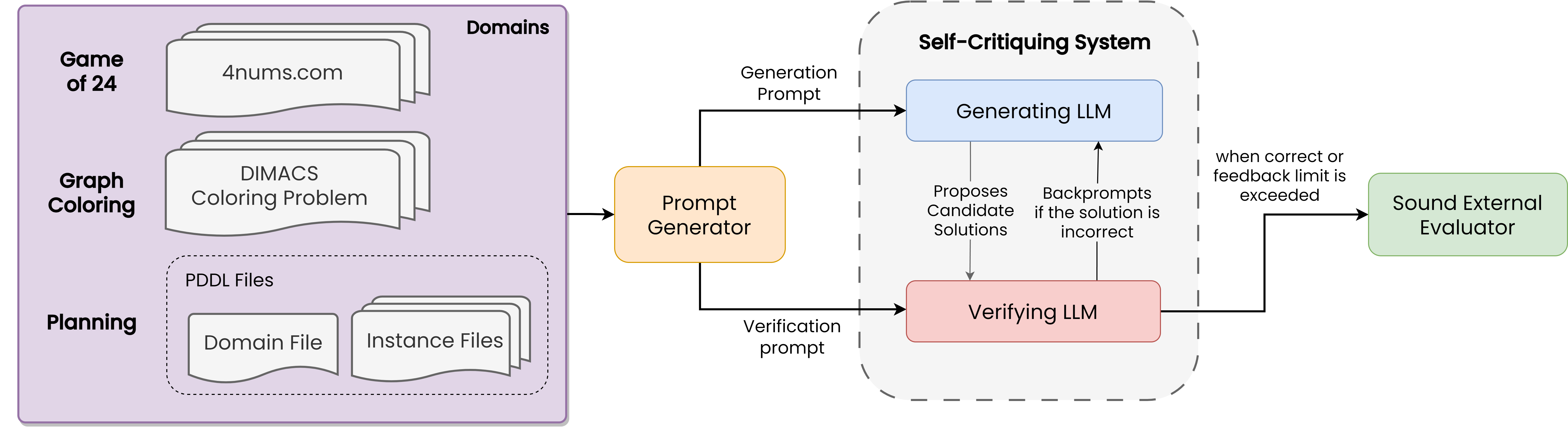}
    \caption{\textbf{Overall Backprompting and Evaluation Architecture}}
    \label{fig:arch}
\end{figure*}

\subsection{STRIPS planning}
STRIPS planning is a formalism used in automated planning that operates in a discrete, deterministic space. Such problems, commonly known as classical planning problems, involve finding a sequence of actions that when executed in a certain world state will take the agent to a desired goal state. STRIPS planning has a long history of featuring in various planning challenges and competitions, and is formally represented using the PDDL (Planning Domain and Definition Language) framework \cite{McDermott1998PDDLthePD}. We consider two domains in STRIPS planning. One is \emph{Blocksworld}, a simple common-sense domain used in International Planning Competitions \cite{ipc} and \emph{Mystery Blocksworld}, which is an obfuscated version of Blocksworld. For both the domains, we draw instances from \cite{valmeekam2023planbench} for our evaluations. 

A PDDL specification consists of three components. The \emph{domain} doesn't change between problems and consists of a set of predicates, which can be used to describe the state of the world, and a set of actions--stored with their preconditions and effects--that the agent is allowed to take. The \emph{initial state} is a list of predicates that are true at the outset of the specific problem (an example predicate, in natural language: "the red block is on the table"). The \emph{goal} is a boolean expression of predicates. 

Solutions to PDDL problems take the form of correct plans--sequences of actions which can be executed from the initial state without violating any of their preconditions and which arrive at a final state that satisfies the goal. Verifying proposed plans is a matter of following the actions in order and checking that these two desiderata are achieved. For our experiments, we use VAL \cite{howey2004val} as the sound external critique that evaluates and critiques LLM generated plans.

\section{Methodology}
As our results are about ablations of self-critique architectures, our basic test framework is a general prompting architecture informed by interchangeable domain-specific components. Our intent is to give the LLM as much information as possible, so we design prompts that include the entire history of previously proposed solutions and the feedback they received.

A problem instance is a domain-specific formal description. In attempting a problem, our system (as shown in Figure \ref{fig:arch}) proceeds as follows: (1) The instance is processed by a simple, hard-coded natural language translator into a prompt which is then sent to the LLM. (2) The LLM's response is wrapped in a domain-specific critique prompt, which is separately sent as another LLM query. (3) If the following response claims that the proposed solution is correct, we stop the system and output the proposed solution. If it doesn't, the critique is extracted, wrapped in instruction text, and appended to a prompt containing the entire history of interactions so far. This is then sent to the LLM, and the cycle repeats, potentially until we enforce a timeout.

Though only two types of prompts are sent, the LLM can be seen as playing three separate roles: as an answer guesser, a (binary) verifier, and a critique generator. In order to better understand which of these roles contribute to increased performance, we will examine variations of this system where one or more of them are changed or removed.

To examine LLM verification abilities, we first measure the performance of the entire system, and then evaluate false positive and false negative rates across domains.
To better understand the guesser role, and the LLM's ability to consider and implement critique, we will modify the loop so that the verification and critique roles are played by a provably sound verifier that provides rich, correct feedback. We will then reduce and eventually eliminate the amount of provided information (e.g. rich feedback: explicitly giving an evaluation of a proposed Game of 24 expression; minimal feedback: "the previous answer was wrong"; no feedback: re-querying with the base prompt), while keeping track of changes in the performance of the entire system. 

For LLM critique generation, we construct subdomains of our original domains. In these prompts, we provide a problem description and a proposed solution, and we ask the LLM to provide domain-specific critique of the solution if it is incorrect. We parse the output using a hard-coded script and measure accuracy compared to the sound verifier's output.

Note that sound verifiers output task specific critiques: for Game of 24, the evaluation of the provided expression (``1+1+4+6=12 not 24"); constraint violations for graph coloring (``vertices 1 and 3 were both colored red despite sharing an edge"); and precondition violations (the second action ``succumb object a" is invalid because the succumb action requires the pain object to be true, which is not the case after the first action.) and failure to reach goal (``this plan does not reach the goal") for planning.

\section{Examining Self-Verification}
\begin{table*}[ht]
    \centering
    \begin{tabular}{p{1.8cm}P{1.2cm}P{1.8cm}cccccP{1.2cm}}
    \toprule
        \textbf{Domain} & \textbf{S.P.} & \textbf{LLM+LLM}  & \multicolumn{3}{c}{
        \centering
        \textbf{LLM+Sound Critique}} & \multicolumn{2}{c}{
        \centering
        \textbf{Sampling}}  & \textbf{S.C.} \\ \cmidrule{4-9}
        &  & & B.F. & F.E.F & A.E.F  & k=15 & k=25 & k=15 \\ \midrule
        Game of 24                       & 5\%                                           & 3\%                                      & 36\%                     & 38\%                    & N/A                            & 28\% & 42\% & 6\%                                \\ \midrule
        Graph Coloring                   & 16\%                                          & 2\%                                 & 38\%               & 37\%               & 34\%                      & 40\%    &  44\%  & 14\%                    \\ \midrule
       Blocksworld                      & 40\%                                            &55\%                                 & 60\%                & 87\%               & 83\% &                      68\% & 72\%  & 42\%                              \\ \midrule
        Mystery Blocksworld              & 4\%                                            & 0\%                                 & 10\%                & 8\%               & 6\%                      & 9\%         & 14\%  & 4\%                     \\ \midrule
         \hline
    \end{tabular}
    \caption{\textbf{Accuracy across prompting schemes over 100 instances per domain.} S.P.-Standard Prompting.
    B.F.-Binary Feedback. F.E.F-First Error Feedback, e.g. the first wrong edge, the first mistaken action, or the non-24 evaluation of the proposed expression. A.E.F-All Error Feedback, e.g. every wrong edge, every mistaken action and error. Note that there is no third critique type for Game of 24 due to the simplicity of the domain.
    We include two examples of sampling, one at 15 samples, the other at 25, to show that completely ablating critique retains the performance increases of critique. We also include S.C.-Self Consistency results, where the most common answer in a pool of 15 is the one that is output by the model, as another comparison point.
    }
    \label{tab:main_results}
\end{table*}
\begin{figure*}
    \centering
    \includegraphics[width=\textwidth]{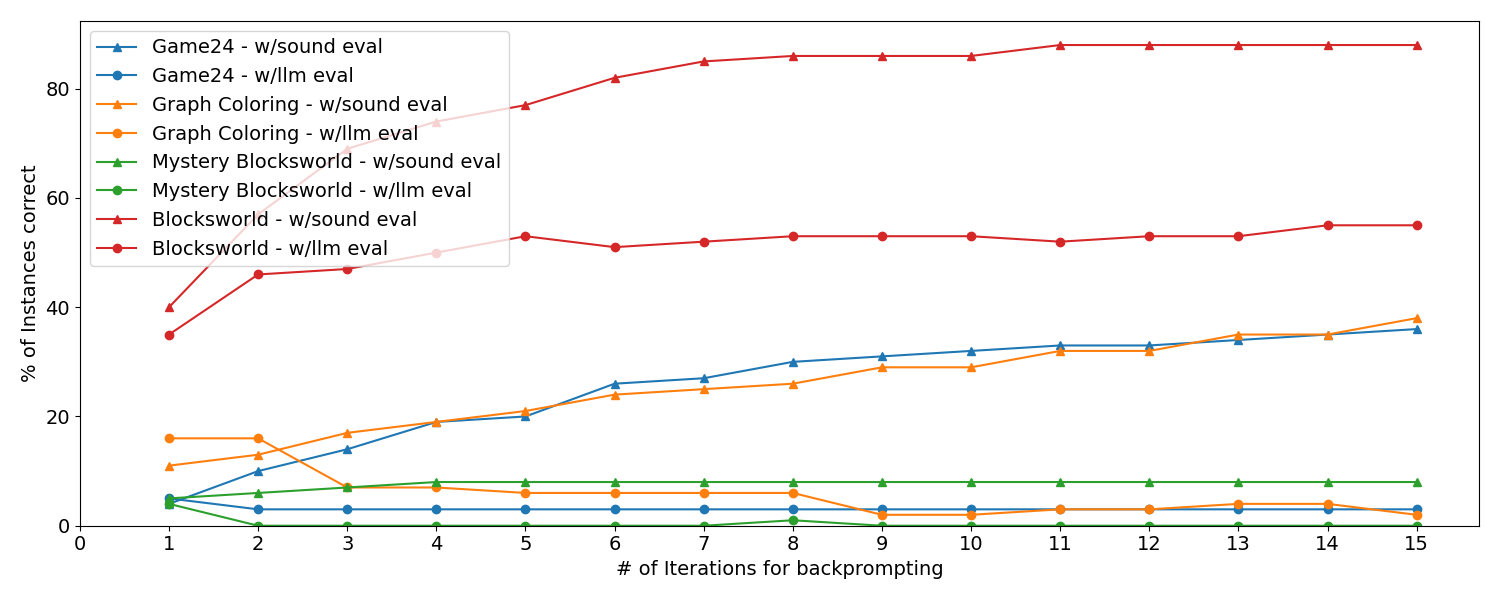}
    \caption{\textbf{Performance vs Number of Iterations Before Timeout.} We measure performance at iteration $n$ by evaluating how many instances are currently correctly solved (whether the LLM has verified them or not. In other words, we evaluate as if the timeout were $n$) and adding that to the number the LLM has correctly verified so far. Note that if the verifier incorrectly rejects an answer and the followup is wrong, the next round may be worse. When paired with a sound verifier, the LLM monotonically improves its performance as the number of backprompts increase up to some asymptote. The top three lines show this for each of our domains. On the other hand, when the LLM itself is used as a verifier, performance collapses immediately.}
    \label{graph:comparison}
\end{figure*}

We evaluate our system over 100 instances in each domain.
In standard prompting we send a single query to the LLM and treat whatever it outputs as its final answer. We use this as our baseline.
As shown in Table \ref{tab:main_results}, when we augment this condition with the full self-critique setup, performance \emph{decreases}.
In fact, Figure \ref{graph:comparison} shows that as the number of backprompts increases, this kind of self-correction consistently degrades output quality. 
If the LLM were a good verifier, then we would have expected it to recognize instances which are already right, and thus--at worst--maintain the baseline score.

The LLM-as-verifier ranges in accuracy depending on the domain, as illustrated in Table \ref{tab:verification_results}.
Notably, Game of 24 and Blocksworld maintain lower rates of both false positives and false negatives, and this is reflected in LLM+LLM performance on those domains, which doesn't fail as drastically as it does in the other cases. In Blocksworld, we even see a modest improvement, though that improvement is still significantly worse than having a sound verifier.
In the remaining two domains, the false negative rates are very high. In effect, the system rejects most answers and then times out on a set of later, worse generations.

When we replace the LLM verifier with a sound verifier, every correct answer will be accepted properly. Intuitively, it can do no worse than standard prompting—anything that was generated correctly initially must be accepted. As shown in Table~\ref{tab:main_results}, performance is much higher in all sound critique cases, though it still falls short of 100\%. Due to the setup, this can't be due to the verifier, but must be the fault of the answer generating LLM. After $15$ rounds, any instance that has yet to been answered correctly will time out, and this process is the \emph{only} source of inaccuracy arising from the LLM-sound verifier loop. 

In general, it is clear that the verifier requires high accuracy or else the overall system will encounter compounding errors. In the reasoning domains considered, LLMs-as-verifiers are mostly insufficient to prevent these.%
\footnote{Prompt and response examples can be found in the Appendix.}

\begin{table*}[ht]
    \centering
    \begin{tabular}{lccc}
    \toprule
        \textbf{Domain}                  & \textbf{Accuracy}            & \textbf{F.P.R}     & \textbf{F.N.R }
          \\ \midrule
        Game of 24                       & 87.0\% (3567/4100)           & 10.4\% (320/3071)  & 20.7\% (213/1029)      
        \\ \midrule

        Graph Coloring                   & 72.4\% (362/500)             &  6.5\% (25/382)    & 95.8\% (113/118)         
        \\ \midrule
        Blocksworld              & 71.8\% (359/500)             & 18.55\% (64/345)      &      15.48\% (24/155) \\ \midrule
        Mystery Blocksworld              & 79.6\% (398/500)             & 0.5\% (2/397)      &      97.09\% (100/103)                   
        \\ \midrule
         \hline
    \end{tabular}
    \caption{\textbf{LLM Verification results.} F.P.R. - False Positive Rate, F.N.R - False Negative Rate.}
    \label{tab:verification_results}
\end{table*}

\subsection{Critique generation}

We consider the quality of LLM-generated free-form critiques separately from that of their binary verification signals, and find that they are full of unhelpful hallucinations and mistakes.
To obtain the following results, we ran a further suites of experiments with specially crafted proposed solutions wrapped in verification prompts. The exact breakdown of which types of solutions were generated is available in each domain's appendix. 

In Game of 24, without any further instructions, the LLM tends to output incorrect suggestions for the answer. When prompted to give an evaluation of the proposed expression first, its accuracy varies. In fact, when we restrict ourselves to only looking at verification of equations that are guaranteed to equal 24, and therefore must be correct, it labels $79.1\%$ of them as correct, but evaluates $81.6\%$ of them to 24. That is, there are problems which it evaluates correctly but which it still marks as wrong. %

In Graph Coloring, the LLM's critiques of proposed solutions are riddled with non-existent edges and include many spurious claims about the colors of vertices, often missing the violated constraint in favor of them. A breakdown and detailed examples are provided in appendix \ref{appendix:coloring_hallucinations}.

In the planning domains, the critiquing LLM often hallucinates whether action preconditions are met or not. In Mystery Blocksworld, the LLM incorrectly assumes the state of these preconditions as well. This leads to lower accuracy of the critiques provided by the LLM. A further breakdown is in appendix~\ref{appendix:mw-critique}

In other words, the LLM introduces errors in two places: verification, where it can pass over correct answers and accept wrong ones; and critique generation, where it can produce misleading feedback and bias future outputs away from the correct answer. When they compound sufficiently, these errors actually reduce the performance of the LLM-based self-critique loop below that of just taking the LLM's very first guess.

\subsection{Critique consideration}

Our results also imply that the LLM often isn't sensitive to varying levels of feedback. We use a sound verifier to critique the LLM's output, and compare the results over three levels of feedback for graph coloring and planning, and over two levels for game of 24. Examples of prompts containing each sort of feedback can be found in appendices \ref{appendix:g24-backprompts}, \ref{appendix:color-backprompts}, and \ref{appendix:mb-backprompts}.
\begin{itemize}
    \item \textbf{Binary feedback} is the same for all domains: either the verifier accepts the solution, stopping the system, or we create a backprompt which says the previous answer was wrong but doesn't elaborate.
    \item \textbf{First error feedback} mentions the first error that was found (e.g. an incorrect edge in graph coloring, an inexecutable step in planning, the evaluation of the proposed expression in game of 24).
    \item \textbf{All errors feedback} includes every error that was found. Note that due to the simplicity of game of 24, we do not implement a third feedback level for it.
\end{itemize}

Perhaps surprisingly, Table \ref{tab:main_results} shows very little difference between these three conditions.
And in two of our domains, increasing the amount of feedback actually leads to a decrease in performance.

The intuition underlying the entire critique system is that sending a history of previous failed attempts together with information about why those attempts were failures should guide the LLM toward better future answers.
If this were true, then we would expect the performance jump to be tied to the content of the backprompts.
With only the data discussed so far, it might seem like the relevant content is actually the history of failed attempts the LLM receives rather than any feedback on those attempts. However, our final experiments contradict this interpretation as well.

We take our ablation of critique consideration to the logical extreme, and remove the availability of critique entirely.
In this sampling setup, we keep the verifier but don't change the prompt at all between iterations.
The LLM (at $t=1$) is asked the same question over and over until the verifier certifies it or it hits some pre-established timeout.

Represented in Table \ref{tab:main_results} by the ``Sampling" columns, this gives comparable gains to feedback conditions. Note that, because prompts do not grow additively with iteration number, the token cost of these prompts is quadratically lower. This allows us to increase performance further by just increasing $k$ further. As a sanity check, we compare this to a self-consistency baseline \cite{wang2022self}, where we instead select the most common answer from the 15 generated ones. This baseline is listed under ``S.C." and shows no improvement over standard prompting.

Our final results show that, in our domains, the information in critiques does not have as much of an effect on performance as previous literature claimed it should. In fact, our performance increases seem to stem in large part just from having enough guesses and a sound verifier. We therefore see the LLM primarily as an idea generator.

\section{Conclusion}

In this paper, we conducted a systematic evaluation of the self-critique abilities of Large Language Models on three reasoning and planning tasks. 
We separated self-critique into three components: verification, critique generation, and critique consideration. Across the hard reasoning domains we evaluated, LLMs did poorly in all three roles, with the stacked errors often making the LLM self-critiquing loop perform worse than just having the LLM guesse the solution up front.  
These failures of verification could potentially be very detrimental to a system's dependability, especially in domains where high reliability is paramount.
In contrast, we saw performance gains when an external sound verifier 
provides the verification signal and critique. 
We also found that good performance can be achieved without any critique whatsoever: just let the LLM make many guesses, and have a sound verifier pick any that is actually correct.

Our results contradict earlier work that has been very optimistic about LLM self-critique abilities. They also add depth to contemporaneous studies that focused on benchmarks that were too easy for LLMs to begin with, lacked clear ground truth, and didn't account for test set memorization. 

Our proposal, based on the case studies we've performed in this paper is, when possible, to embed LLMs in systems which allow them to guess at solutions multiple times, but which provide some kind of signal for when a guess is good enough.
Ideally, this takes the form of a sound verifier, like VAL \cite{howey2004val} for STRIPS planning, basic expression simplification for Game of 24, or a constraint checker for constraint satisfaction problems.
In real-world applications we expect this role to be played by a menagerie of partial critics evaluating plans or solutions based on criteria that they have access to, designed so that consensus is considered verification. Similar architectures have already shown some success \cite{funsearch}, and previous work has proposed the general LLM-Modulo framework \cite{kambhampati2024llms} which the current work fits into.

\bibliography{llmplan}

\begin{thebibliography}{10}

\bibitem{abdin2023kitab}
Marah~I Abdin, Suriya Gunasekar, Varun Chandrasekaran, Jerry Li, Mert Yuksekgonul, Rahee~Ghosh Peshawaria, Ranjita Naik, and Besmira Nushi.
\newblock Kitab: Evaluating llms on constraint satisfaction for information retrieval.
\newblock {\em arXiv preprint arXiv:2310.15511}, 2023.

\bibitem{achiam2023gpt}
Josh Achiam, Steven Adler, Sandhini Agarwal, Lama Ahmad, Ilge Akkaya, Florencia~Leoni Aleman, Diogo Almeida, Janko Altenschmidt, Sam Altman, Shyamal Anadkat, et~al.
\newblock Gpt-4 technical report.
\newblock {\em arXiv preprint arXiv:2303.08774}, 2023.

\bibitem{arakelyan2023exploring}
Shushan Arakelyan, Rocktim~Jyoti Das, Yi~Mao, and Xiang Ren.
\newblock Exploring distributional shifts in large language models for code analysis.
\newblock {\em arXiv preprint arXiv:2303.09128}, 2023.

\bibitem{bubeck2023sparks}
S{\'e}bastien Bubeck, Varun Chandrasekaran, Ronen Eldan, Johannes Gehrke, Eric Horvitz, Ece Kamar, Peter Lee, Yin~Tat Lee, Yuanzhi Li, Scott Lundberg, et~al.
\newblock Sparks of artificial general intelligence: Early experiments with gpt-4.
\newblock {\em arXiv preprint arXiv:2303.12712}, 2023.

\bibitem{chen2023teaching}
Xinyun Chen, Maxwell Lin, Nathanael Sch{\"a}rli, and Denny Zhou.
\newblock Teaching large language models to self-debug.
\newblock {\em arXiv preprint arXiv:2304.05128}, 2023.

\bibitem{cohen2023lm}
Roi Cohen, May Hamri, Mor Geva, and Amir Globerson.
\newblock Lm vs lm: Detecting factual errors via cross examination.
\newblock {\em arXiv preprint arXiv:2305.13281}, 2023.

\bibitem{de2023can}
Joost~CF de~Winter.
\newblock Can chatgpt pass high school exams on english language comprehension?
\newblock {\em International Journal of Artificial Intelligence in Education}, pages 1--16, 2023.

\bibitem{dimacsrutgers}
DIMACS.
\newblock {DIMACS} {Implementation} {Challenges}.
\newblock Archive available at \url{http://archive.dimacs.rutgers.edu/Challenges/}.

\bibitem{dong2022survey}
Qingxiu Dong, Lei Li, Damai Dai, Ce~Zheng, Zhiyong Wu, Baobao Chang, Xu~Sun, Jingjing Xu, and Zhifang Sui.
\newblock A survey on in-context learning.
\newblock {\em arXiv preprint arXiv:2301.00234}, 2022.

\bibitem{du2023improving}
Yilun Du, Shuang Li, Antonio Torralba, Joshua~B Tenenbaum, and Igor Mordatch.
\newblock Improving factuality and reasoning in language models through multiagent debate.
\newblock {\em arXiv preprint arXiv:2305.14325}, 2023.

\bibitem{dziri2023faith}
Nouha Dziri, Ximing Lu, Melanie Sclar, Xiang~Lorraine Li, Liwei Jiang, Bill~Yuchen Lin, Sean Welleck, Peter West, Chandra Bhagavatula, Ronan~Le Bras, Jena~D. Hwang, Soumya Sanyal, Xiang Ren, Allyson Ettinger, Zaid Harchaoui, and Yejin Choi.
\newblock Faith and fate: Limits of transformers on compositionality.
\newblock In {\em Thirty-seventh Conference on Neural Information Processing Systems}, 2023.

\bibitem{gendron2023large}
Ga{\"e}l Gendron, Qiming Bao, Michael Witbrock, and Gillian Dobbie.
\newblock Large language models are not abstract reasoners.
\newblock {\em arXiv preprint arXiv:2305.19555}, 2023.

\bibitem{gilson2023does}
Aidan Gilson, Conrad~W Safranek, Thomas Huang, Vimig Socrates, Ling Chi, Richard~Andrew Taylor, David Chartash, et~al.
\newblock How does chatgpt perform on the united states medical licensing examination (usmle)? the implications of large language models for medical education and knowledge assessment.
\newblock {\em JMIR medical education}, 9(1):e45312, 2023.

\bibitem{gou2023critic}
Zhibin Gou, Zhihong Shao, Yeyun Gong, Yelong Shen, Yujiu Yang, Nan Duan, and Weizhu Chen.
\newblock Critic: Large language models can self-correct with tool-interactive critiquing.
\newblock {\em arXiv preprint arXiv:2305.11738}, 2023.

\bibitem{hao_reasoning_2023}
Shibo Hao, Yi~Gu, Haodi Ma, Joshua Hong, Zhen Wang, Daisy Wang, and Zhiting Hu.
\newblock Reasoning with {Language} {Model} is {Planning} with {World} {Model}.
\newblock In Houda Bouamor, Juan Pino, and Kalika Bali, editors, {\em Proceedings of the 2023 {Conference} on {Empirical} {Methods} in {Natural} {Language} {Processing}}, pages 8154--8173, Singapore, December 2023. Association for Computational Linguistics.

\bibitem{hong2023closer}
Ruixin Hong, Hongming Zhang, Xinyu Pang, Dong Yu, and Changshui Zhang.
\newblock A closer look at the self-verification abilities of large language models in logical reasoning.
\newblock {\em arXiv preprint arXiv:2311.07954}, 2023.

\bibitem{howey2004val}
Richard Howey, Derek Long, and Maria Fox.
\newblock {VAL: Automatic plan validation, continuous effects and mixed initiative planning using PDDL}.
\newblock In {\em 16th IEEE International Conference on Tools with Artificial Intelligence}, pages 294--301. IEEE, 2004.

\bibitem{huang2023large}
Jie Huang, Xinyun Chen, Swaroop Mishra, Huaixiu~Steven Zheng, Adams~Wei Yu, Xinying Song, and Denny Zhou.
\newblock Large language models cannot self-correct reasoning yet.
\newblock {\em arXiv preprint arXiv:2310.01798}, 2023.

\bibitem{huang2022inner}
Wenlong Huang, Fei Xia, Ted Xiao, Harris Chan, Jacky Liang, Pete Florence, Andy Zeng, Jonathan Tompson, Igor Mordatch, Yevgen Chebotar, et~al.
\newblock Inner monologue: Embodied reasoning through planning with language models.
\newblock {\em arXiv preprint arXiv:2207.05608}, 2022.

\bibitem{ipc}
IPC.
\newblock International planning competition, 1998.

\bibitem{jiang2023backward}
Weisen Jiang, Han Shi, Longhui Yu, Zhengying Liu, Yu~Zhang, Zhenguo Li, and James~T Kwok.
\newblock Backward reasoning in large language models for verification.
\newblock {\em arXiv preprint arXiv:2308.07758}, 2023.

\bibitem{kambhampati2024can}
Subbarao Kambhampati.
\newblock Can large language models reason and plan?
\newblock {\em Annals of the New York Academy of Sciences}, 2024.

\bibitem{kambhampati2024llms}
Subbarao Kambhampati, Karthik Valmeekam, Lin Guan, Kaya Stechly, Mudit Verma, Siddhant Bhambri, Lucas Saldyt, and Anil Murthy.
\newblock Llms can't plan, but can help planning in llm-modulo frameworks.
\newblock {\em arXiv preprint arXiv:2402.01817}, 2024.

\bibitem{kambhampatirole}
Subbarao Kambhampati, Karthik Valmeekam, Matthew Marquez, and Lin Guan.
\newblock On the role of large language models in planning. tutorial presented at the international conference on automated planning and scheduling (icaps), prague, July 2023.

\bibitem{kojima2022large}
Takeshi Kojima, Shixiang~Shane Gu, Machel Reid, Yutaka Matsuo, and Yusuke Iwasawa.
\newblock Large language models are zero-shot reasoners.
\newblock {\em Advances in neural information processing systems}, 35:22199--22213, 2022.

\bibitem{kortemeyer2023could}
Gerd Kortemeyer.
\newblock Could an artificial-intelligence agent pass an introductory physics course?
\newblock {\em Physical Review Physics Education Research}, 19(1):010132, 2023.

\bibitem{leighton2004nature}
Jacqueline~P Leighton and Robert~J Sternberg.
\newblock {\em The nature of reasoning}.
\newblock Cambridge University Press, 2004.

\bibitem{lin2020commongen}
Bill~Yuchen Lin, Wangchunshu Zhou, Ming Shen, Pei Zhou, Chandra Bhagavatula, Yejin Choi, and Xiang Ren.
\newblock Commongen: A constrained text generation challenge for generative commonsense reasoning.
\newblock In {\em Findings of the Association for Computational Linguistics: EMNLP 2020}, pages 1823--1840, 2020.

\bibitem{madaan2023self}
Aman Madaan, Niket Tandon, Prakhar Gupta, Skyler Hallinan, Luyu Gao, Sarah Wiegreffe, Uri Alon, Nouha Dziri, Shrimai Prabhumoye, Yiming Yang, et~al.
\newblock Self-refine: Iterative refinement with self-feedback.
\newblock {\em arXiv preprint arXiv:2303.17651}, 2023.

\bibitem{mccoy2023embers}
R~Thomas McCoy, Shunyu Yao, Dan Friedman, Matthew Hardy, and Thomas~L Griffiths.
\newblock Embers of autoregression: Understanding large language models through the problem they are trained to solve.
\newblock {\em arXiv preprint arXiv:2309.13638}, 2023.

\bibitem{McDermott1998PDDLthePD}
Drew McDermott, Malik Ghallab, Adele~E. Howe, Craig~A. Knoblock, Ashwin Ram, Manuela~M. Veloso, Daniel~S. Weld, and David~E. Wilkins.
\newblock Pddl-the planning domain definition language.
\newblock 1998.

\bibitem{rae2021scaling}
Jack~W Rae, Sebastian Borgeaud, Trevor Cai, Katie Millican, Jordan Hoffmann, Francis Song, John Aslanides, Sarah Henderson, Roman Ring, Susannah Young, et~al.
\newblock Scaling language models: Methods, analysis \& insights from training gopher.
\newblock {\em arXiv preprint arXiv:2112.11446}, 2021.

\bibitem{raimondi2023comparative}
Raffaele Raimondi, Nikolaos Tzoumas, Thomas Salisbury, Sandro Di~Simplicio, and Mario~R Romano.
\newblock Comparative analysis of large language models in the royal college of ophthalmologists fellowship exams.
\newblock {\em Eye}, 37(17):3530--3533, 2023.

\bibitem{roberts2023data}
Manley Roberts, Himanshu Thakur, Christine Herlihy, Colin White, and Samuel Dooley.
\newblock Data contamination through the lens of time.
\newblock {\em arXiv preprint arXiv:2310.10628}, 2023.

\bibitem{funsearch}
Bernardino Romera-Paredes, Mohammadamin Barekatain, Alexander Novikov, Matej Balog, M~Pawan Kumar, Emilien Dupont, Francisco~JR Ruiz, Jordan~S Ellenberg, Pengming Wang, Omar Fawzi, et~al.
\newblock Mathematical discoveries from program search with large language models.
\newblock {\em Nature}, pages 1--3, 2023.

\bibitem{saxon_benchmarks_2024}
Michael Saxon.
\newblock Benchmarks as {Microscopes}: {A} {Call} for {Model} {Metrology}, 2024.

\bibitem{shinn2023reflexion}
Noah Shinn, Federico Cassano, Ashwin Gopinath, Karthik~R Narasimhan, and Shunyu Yao.
\newblock Reflexion: Language agents with verbal reinforcement learning.
\newblock In {\em Thirty-seventh Conference on Neural Information Processing Systems}, 2023.

\bibitem{silver2022pddl}
Tom Silver, Varun Hariprasad, Reece~S Shuttleworth, Nishanth Kumar, Tom{\'a}s Lozano-P{\'e}rez, and Leslie~Pack Kaelbling.
\newblock Pddl planning with pretrained large language models.
\newblock In {\em NeurIPS 2022 foundation models for decision making workshop}, 2022.

\bibitem{srivastava2023beyond}
Aarohi Srivastava, Abhinav Rastogi, Abhishek Rao, Abu Awal~Md Shoeb, Abubakar Abid, Adam Fisch, Adam~R Brown, Adam Santoro, Aditya Gupta, Adri{\`a} Garriga-Alonso, et~al.
\newblock Beyond the imitation game: Quantifying and extrapolating the capabilities of language models.
\newblock {\em Transactions on Machine Learning Research}, 2023.

\bibitem{stechly2024chain}
Kaya Stechly, Karthik Valmeekam, and Subbarao Kambhampati.
\newblock Chain of thoughtlessness: An analysis of cot in planning.
\newblock {\em arXiv preprint arXiv:2405.04776}, 2024.

\bibitem{thaker2024large}
Nikhil~G Thaker, Navid Redjal, Arturo Loaiza-Bonilla, David Penberthy, Tim Showalter, Ajay Choudhri, Shirnett Williamson, Gautam Thaker, Chirag Shah, Matthew~C Ward, et~al.
\newblock Large language models encode radiation oncology domain knowledge: Performance on the american college of radiology standardized examination.
\newblock {\em AI in Precision Oncology}, 1(1):43--50, 2024.

\bibitem{ullman2023large}
Tomer Ullman.
\newblock Large language models fail on trivial alterations to theory-of-mind tasks.
\newblock {\em arXiv preprint arXiv:2302.08399}, 2023.

\bibitem{valmeekam2023planbench}
Karthik Valmeekam, Matthew Marquez, Alberto Olmo, Sarath Sreedharan, and Subbarao Kambhampati.
\newblock Planbench: An extensible benchmark for evaluating large language models on planning and reasoning about change.
\newblock In {\em Thirty-seventh Conference on Neural Information Processing Systems Datasets and Benchmarks Track}, 2023.

\bibitem{valmeekam2023on}
Karthik Valmeekam, Matthew Marquez, Sarath Sreedharan, and Subbarao Kambhampati.
\newblock On the planning abilities of large language models - a critical investigation.
\newblock In {\em Thirty-seventh Conference on Neural Information Processing Systems}, 2023.

\bibitem{wang2023scibench}
Xiaoxuan Wang, Ziniu Hu, Pan Lu, Yanqiao Zhu, Jieyu Zhang, Satyen Subramaniam, Arjun~R Loomba, Shichang Zhang, Yizhou Sun, and Wei Wang.
\newblock Scibench: Evaluating college-level scientific problem-solving abilities of large language models.
\newblock {\em arXiv preprint arXiv:2307.10635}, 2023.

\bibitem{wang2022self}
Xuezhi Wang, Jason Wei, Dale Schuurmans, Quoc Le, Ed~Chi, Sharan Narang, Aakanksha Chowdhery, and Denny Zhou.
\newblock Self-consistency improves chain of thought reasoning in language models.
\newblock {\em arXiv preprint arXiv:2203.11171}, 2022.

\bibitem{webb2023emergent}
Taylor Webb, Keith~J Holyoak, and Hongjing Lu.
\newblock Emergent analogical reasoning in large language models.
\newblock {\em Nature Human Behaviour}, 7(9):1526--1541, 2023.

\bibitem{wei2022chain}
Jason Wei, Xuezhi Wang, Dale Schuurmans, Maarten Bosma, Fei Xia, Ed~Chi, Quoc~V Le, Denny Zhou, et~al.
\newblock Chain-of-thought prompting elicits reasoning in large language models.
\newblock {\em Advances in neural information processing systems}, 35:24824--24837, 2022.

\bibitem{welleck2022generating}
Sean Welleck, Ximing Lu, Peter West, Faeze Brahman, Tianxiao Shen, Daniel Khashabi, and Yejin Choi.
\newblock Generating sequences by learning to self-correct.
\newblock In {\em The Eleventh International Conference on Learning Representations}, 2022.

\bibitem{weng2023large}
Yixuan Weng, Minjun Zhu, Fei Xia, Bin Li, Shizhu He, Shengping Liu, Bin Sun, Kang Liu, and Jun Zhao.
\newblock Large language models are better reasoners with self-verification.
\newblock In {\em Findings of the Association for Computational Linguistics: EMNLP 2023}, pages 2550--2575, 2023.

\bibitem{yang2018hotpotqa}
Zhilin Yang, Peng Qi, Saizheng Zhang, Yoshua Bengio, William Cohen, Ruslan Salakhutdinov, and Christopher~D Manning.
\newblock Hotpotqa: A dataset for diverse, explainable multi-hop question answering.
\newblock In {\em Proceedings of the 2018 Conference on Empirical Methods in Natural Language Processing}, pages 2369--2380, 2018.

\bibitem{yao2023tree}
Shunyu Yao, Dian Yu, Jeffrey Zhao, Izhak Shafran, Thomas~L. Griffiths, Yuan Cao, and Karthik~R Narasimhan.
\newblock Tree of thoughts: Deliberate problem solving with large language models.
\newblock In {\em Thirty-seventh Conference on Neural Information Processing Systems}, 2023.

\bibitem{yao2022react}
Shunyu Yao, Jeffrey Zhao, Dian Yu, Nan Du, Izhak Shafran, Karthik Narasimhan, and Yuan Cao.
\newblock React: Synergizing reasoning and acting in language models.
\newblock {\em arXiv preprint arXiv:2210.03629}, 2022.

\bibitem{yeadon2023exploring}
Will Yeadon and Douglas~P Halliday.
\newblock Exploring durham university physics exams with large language models.
\newblock {\em arXiv preprint arXiv:2306.15609}, 2023.

\bibitem{zhou2022least}
Denny Zhou, Nathanael Sch{\"a}rli, Le~Hou, Jason Wei, Nathan Scales, Xuezhi Wang, Dale Schuurmans, Claire Cui, Olivier Bousquet, Quoc Le, et~al.
\newblock Least-to-most prompting enables complex reasoning in large language models.
\newblock {\em arXiv preprint arXiv:2205.10625}, 2022.

\end{thebibliography}
\bibliographystyle{plain}

\newpage
\appendix
\onecolumn
\section{Appendix}
\setcounter{table}{0}
\renewcommand{\thetable}{A\arabic{table}}

\subsection{Prompt Variation and Chain of Thought}
LLM results are well-known to be brittle to choice and phrasing of prompt. We ran experiments across multiple prompts to ensure the robustness of our results. Table~\ref{tab:fullv-cot} shows the results for the full pipeline where verification prompts are modified to ask for CoT reasoning first. Results from the main paper are provided for easy comparison. Prompts can be found in each domain's prompt section in this appendix, under the header "Prompt to Elicit CoT Verification".

Performance does improve in some, though not all, cases. However, seemingly near-perfect improvements in verification ability do not translate into near-sound-verifier performance. In G24, CoT increases verification accuracy from 87\% to 99\%, and does shrink the difference between sound verifier and LLM-verifier in the full pipeline, but a 6 percentage point gap still remains! Furthermore, these improvements come with a large cost increase: in G24, this leads to a 17 times increase in necessary output tokens, which more than doubles the cost of verification.

Note that Chain of Thought techniques themselves vary greatly in their effectiveness across domains. In particular, in reasoning domains like Blocksworld, previous work has shown that they fail to generalize and are not particularly robust~\cite{stechly2024chain}.

On the rest of the tasks, where verification is also fairly simple and linear, and thus theoretically amenable to CoT, we do not see nearly as significant improvements.

\begin{table*}[ht]
    \centering
    \begin{tabular}{p{1.8cm}P{1.2cm}P{1.8cm}ccP{1.2cm}}
    \toprule
        \textbf{Domain}      & \textbf{S.P.} & \textbf{Sampling} & \textbf{LLM+LLM} & \textbf{LLM+LLM-C} & \textbf{F.E.F.} \\\midrule
        Game of 24           & 5\%           & 28\%              &  3\%             & 32\%               & 38\%            \\ \midrule
        Coloring             & 16\%          & 40\%              &  2\%             &  0\%               & 37\%            \\ \midrule
        Blocksworld          & 40\%          & 68\%              & 55\%             & 53\%               & 87\%            \\ \midrule
        Mystery              & 4\%           &  9\%              &  0\%             &  5\%               &  6\%            \\ \midrule
        \hline
    \end{tabular}
    \caption{\textbf{Accuracy across prompting schemes including CoT verification schemes over 100 instances per domain.} S.P.--Standard Prompting. Sampling--k=15. LLM+LLM--Main paper result. LLM+LLM-C--Full pipeline with chain of thought verification. F.E.F--(Sound) First Error Feedback.}
    \label{tab:fullv-cot}
\end{table*}

We also reran our verification-only experiments with these new prompts, as well as with variations on the original (non-CoT) prompts. Table~\ref{tab:onlyv-cot} has these results, presented alongside the original ones.
\begin{table*}[ht]
    \centering
    \begin{tabular}{lccc}
    \toprule
        \textbf{Domain}          & \textbf{Accuracy}  & \textbf{F.P.R}     & \textbf{F.N.R }   \\\midrule
        Game of 24               & 87.0\% (3567/4100) & 10.4\%  (320/3071) & 20.7\%  (213/1029)\\\midrule
        Game of 24-CoT           & 98.8\% (4051/4100) &  0.2\%  (6/3071)   &  4.3\%  (44/1029) \\\midrule
        Graph Coloring           & 72.4\% (362/500)   &  6.5\%  (25/382)   & 95.8\%  (113/118) \\\midrule
        Graph Coloring-CoT       & 77.6\% (388/500)   & 10.7\%  (41/382)   & 60.2\%  (71/118)  \\\midrule
        Blocksworld              & 71.8\% (359/500)   & 18.55\% (64/345)   & 15.48\% (24/155)  \\\midrule
        Blocksworld-S            & 71.2\% (356/500)   & 22.1\%  (76/345)   &  8.4\%  (13/155)  \\\midrule
        Blocksworld-CoT          & 77.6\% (388/500)   &  7.6\%  (26/345)   & 23.9\%  (37/155)  \\\midrule
        Mystery                  & 79.6\% (398/500)   &  0.5\%  (2/397)    & 97.09\% (100/103) \\\midrule
        Mystery-S                & 79.0\% (395/500)   &  1.3\%  (4/397)    & 96.1\%  (149/103) \\\midrule
        Mystery-CoT              & 81.8\% (409/500)   &  3.2\%  (11/397)   & 72.8\%  (113/103) \\\midrule
         \hline
    \end{tabular}
    \caption{\textbf{LLM Verification results across prompts.} F.P.R. - False Positive Rate, F.N.R - False Negative Rate. The -S cases are non-CoT prompts with the answer and reasoning swapped for those domains where answer was originally asked for first. The -CoT cases are those in which verification is done with CoT.}
    \label{tab:onlyv-cot}
\end{table*}

\subsection{On Tree of Thoughts}
Our results on the Game of 24 setting seem to contradict the results shown in \cite{yao2023tree}. However, this is mainly because the self-verification setting presented in the main text of this paper is not directly comparable to that of \cite{yao2023tree}. We ran an additional analysis to provide a direct comparison. 

Our external verifier results are all done with only 15 queries to the LLM. \cite{yao2023tree} isn’t entirely clear on the number of queries used, but table 7 in the appendix does give a cost breakdown. Per problem, 100 CoT prompts costs \$0.47, but running Tree of Thoughts (ToT) averages \$0.74--cost-equivalent to about 150 CoT prompts. On the exact same test set, we extend our experiments to 150 (direct, non-CoT) queries with a sound verifier, and we reach 70\%, comparable to ToT’s reported 74\%.

The remaining difference is likely due to the fact that ToT implements a classical breadth-first search algorithm, only prompting the LLM to generate (much easier) intermediate steps and heuristic evaluations rather than full solutions and reflections. By reducing compositionality and offloading it to a proven classical algorithm, ToT sidesteps some of the major hurdles to LLM reasoning. Our results highlight why such techniques do not scale beyond the simplest toy instances.

\subsection{Game of 24}
\subsubsection{Evaluation vs. Binary Verification for Game of 24}
The following is a more in-depth comparison of GPT-4's critique and verification abilities on game of 24.

For each instance, we generated five different kinds of proposed expressions: correct, ablated operation (exactly one operation is wrong), ablated number (exactly one number is wrong), random, and LLM (sampled from LLM generations). For each of these proposed expression, we sent a query to the LLM asking it to first evaluate the expression and then to say if it is correct, that is equals 24. We also generated two more "no info" cases: correct and random. These two are the exact same as the previous, but only ask for the evaluation of an expression without mentioning the associated goal state (=24) or asking for verification.

Table \ref{table:g24verification} summarizes the results. Note that we generated 1000 expressions for each type, one from every problem in the full set, but only 100 for the LLM case, as our generations were constrained in the main paper to instances 901-1000.

\begin{table}[]
\centering
\begin{tabular}{lll}
\toprule
\multicolumn{1}{c}{} & Correct Evaluation & Correct Verification \\ \midrule
\multicolumn{1}{l}{correct}         & 81.6\%       & 79.1\%               \\ \midrule
\multicolumn{1}{l}{correct-no-info} & 84.4\%       & -                    \\ \midrule
\multicolumn{1}{l}{ablated\_op}     & 47.5\%       & 92.1\%               \\ \midrule
\multicolumn{1}{l}{ablated\_number} & 52.2\%       & 82.9\%               \\ \midrule
\multicolumn{1}{l}{random}          & 48.8\%       & 95.5\%               \\ \midrule
\multicolumn{1}{l}{random-no-info}  & 60.3\%       & -                    \\ \midrule
\multicolumn{1}{l}{LLM}             & 55\%         & 71\%                 \\ \midrule
\bottomrule
\end{tabular}
\label{table:g24verification}
\caption{GPT-4's evaluation vs. verification on Game of 24 across expression types.}
\end{table}

\subsubsection{Prompts}
All of following examples are built on the same Game of 24 problem, except for the LLM Self-Critique examples.
\mytcbinput{appendix/game24/problem_example.tex}{Raw text format of Game of 24 instance}{0}{g24_example:g24instance}
\mytcbinput{appendix/game24/base_prompt.tex}{Baseline{,} Direct Prompt}{0}{g24_prompt:baseline}
\mytcbinput{appendix/game24/example_response.tex}{Example LLM Response}{0}{g24_example:response}
\mytcbinput{appendix/game24/llm_v_prompt.tex}{Prompt To Elicit Verification}{0}{g24_prompt:verification}
\mytcbinput{appendix/game24/llm_v_prompt_cot.tex}{Prompt To Elicit CoT Verification}{0}{g24_prompt:verification_cot}

\subsubsection{Backprompts}\label{appendix:g24-backprompts}
\mytcbinput{appendix/game24/llm-self_example_backprompt.tex}{Backprompt Generated From Self-Critique}{0}{g24_backprompt:self}
\mytcbinput{appendix/game24/passfail_example_backprompt.tex}{Backprompt Generated From Sound Verifier (Pass/Fail)}{0}{g24_backprompt:passfail}
\mytcbinput{appendix/game24/first_example_backprompt.tex}{Backprompt Generated From Sound Verifier}{0}{g24_backprompt:first}
\mytcbinput{appendix/game24/malformed_example_passfail.tex}{Backprompt Generated From Sound Verifier Where the Answer Was Malformed (Missing a Parenthesis)}{0}{g24_backprompt:full}

\subsubsection{Complete Example of Iterative Prompt Sequence}
\mytcbinput{appendix/game24/full_trace_example.tex}{Full Sequence of 7 (Sound Verifier Generated) Backprompts Stopping at Successful Completion of the Task}{0}{g24_example:trace}

\subsection{Graph Coloring}
\subsubsection{Hallucinations in Graph Coloring Critique}\label{appendix:coloring_hallucinations}

The following is a more in-depth look at GPT-4's critique abilities on the graph coloring problem.

For each instance, we generated five different kinds of colorings: correct, ablated (exactly one edge is wrong), non-optimal (a coloring with no constraint violations except that it uses too many colors), random, and LLM (sampled from LLM generations). For each of these 500 proposed colorings, we sent a query to the LLM asking it to verify if the answer was right, and--if not--to output a free-form critique. We then parsed these to determine what edges it said were colored the same at both vertices.

Table \ref{table:hallucination} summarizes the results. Note that, proportionally, hallucinations decrease when the error rate of the domain increases. That is to say, when there are more incorrect edges, the model is more likely to point to one of them. Intuitively, this makes sense: it's easier to guess one edge which is wrong when half of all the edges are miscolored, as is the case on average among randomly colored graphs.

\begin{table}[]
\caption{Distribution of hallucinations during verification task. This table counts the number of instances that featured each type of hallucination and compares it to the total number of erroneous edges encountered across all coloring instances in each subset. Note that the correct column tracks the number of colorings where the \emph{ground truth} is that the coloring is correct.}
  \label{table:hallucination}
  \centering
\begin{tabular}{@{}lllllll@{}}
\toprule
                                 & \multicolumn{4}{c}{Hallucinations}               & \multicolumn{2}{c}{Coloring} \\ \midrule
\multicolumn{1}{l|}{}            & Vertex & Edge & Both & \multicolumn{1}{l|}{None} & Errors       & Correct       \\
\multicolumn{1}{l|}{Correct}     & 29     & 72   & 7    & \multicolumn{1}{l|}{2}    & 0            & 100           \\
\multicolumn{1}{l|}{Ablated}     & 24     & 52   & 5    & \multicolumn{1}{l|}{24}   & 187          & 0             \\
\multicolumn{1}{l|}{Non-optimal} & 18     & 65   & 3    & \multicolumn{1}{l|}{10}   & 0            & 0             \\
\multicolumn{1}{l|}{Random}      & 10     & 26   & 5    & \multicolumn{1}{l|}{66}   & 736          & 0             \\
\multicolumn{1}{l|}{LLM}         & 26     & 41   & 6    & \multicolumn{1}{l|}{27}   & 240          & 18            \\ \midrule
\multicolumn{1}{l|}{Total}       & 107    & 256  & 26   & \multicolumn{1}{l|}{129}  & 282          & 118
\end{tabular}
\end{table}

Edge hallucinations are more common than vertex. Essentially, typical behavior is to pick two vertices that are the same color in the coloring, but which aren't associated by an edge in the graph description, and claim that they are connected and thus illegally colored. Vertex color hallucination is when the reverse happens: instead of ascribing an edge to same-color nodes, the colorings of two connected vertices are misstated. The overlap between the two cases, where a non-existent edge is declared to be violated by non-existent colorings is much rarer than either. Note that it never hallucinates new vertex \emph{names}, only that vertices which are in graph have colors differing from reality.

Even rarer cases did spring up in the response data. At times the model lost track of the question being asked and reversed it, explicitly claiming that two same-colored vertices violate the conditions because they \emph{aren't} connected; or it began to contradict itself mid-deduction, making multiple claims about a vertex's color.

We present these examples here. In the following, multiple equals signs in a row ("===") are dividers between separate examples, not parts of the examples.
\mytcbinput{appendix/csp/LLM_V_example_normal.tex}{Examples of LLM Output on the Verification Task}{0}{gc_example:normal_v_responses}
\mytcbinput{appendix/csp/LLM_V_example_weird.tex}{Examples of (Rare) Mangled{,} Illogical{,} and Otherwise Strange Output on the Verification Task}{0}{gc_example:weird_v_responses}

\subsubsection{Prompts}
All of following examples are built on the same graph instance.
\mytcbinput{appendix/csp/dimacs_example.tex}{DIMACS Format For Graphs}{0}{gc_example:dimacs}
\mytcbinput{appendix/csp/base_prompt.tex}{Baseline{,} Direct Prompt}{0}{gc_prompt:baseline}
\mytcbinput{appendix/csp/example_response.tex}{Example LLM Response}{0}{gc_example:response}
\mytcbinput{appendix/csp/llm_v_prompt.tex}{Prompt To Elicit Verification}{0}{gc_prompt:verification}
\mytcbinput{appendix/csp/llm_v_prompt_cot.tex}{Prompt To Elicit CoT Verification}{0}{gc_prompt:verification_cot}

\subsubsection{Backprompts} \label{appendix:color-backprompts}
\mytcbinput{appendix/csp/llm-self_example_backprompt.tex}{Backprompt Generated From Self-Critique}{0}{gc_backprompt:self}
\mytcbinput{appendix/csp/passfail_example_backprompt.tex}{Backprompt Generated From Sound Verifier (Pass/Fail)}{0}{gc_backprompt:passfail}
\mytcbinput{appendix/csp/first_example_backprompt.tex}{Backprompt Generated From Sound Verifier (First)}{0}{gc_backprompt:first}
\mytcbinput{appendix/csp/full_example_backprompt.tex}{Backprompt Generated From Sound Verifier (Full)}{0}{gc_backprompt:full}

\subsubsection{Complete Example of Iterative Prompt Sequence}
\mytcbinput{appendix/csp/full_trace_example.tex}{Full Sequence of 15 (LLM-Generated) Backprompts}{0}{gc_example:trace}

\subsection{Planning}
\subsubsection{Accuracy of Planning Critique}\label{appendix:mw-critique}
For each instance, we generated five different kind of plans: correct, inexecutable (where an action is inexecutable and the plan is thus invalid), non goal reaching (all actions are executable but the plan does not reach the desired state), random, and LLM (sampled from the LLM generations). For each of these prompts, we sent a query to the LLM asking it to verify the plan and critique it in a certain format. We provide an example in the following prompts section. Specifically, if the plan is valid, the LLM has to just output that. If a plan is invalid and inexecutable, it has to provide the first action that the plan is rendered invalid and the unmet preconditions for that action. If the plan is invalid and non goal reaching, it has to provide the unmet goals for the plan. We evaluate the binary verification and the critique by comparing it to the ground truth provided by VAL \cite{howey2004val}. We check the binary verification, whether or not the detected the type of invalidity (inexecutable or non goal reaching) and if the critique is valid. Tables \ref{tab:blocksworld} and \ref{tab:mystery} show the inability of LLMs in generating the right critique for proposed solutions even though they get the label correct.

\begin{table}[]
    \centering
    \begin{tabular}{lllll}
    \toprule
         \multicolumn{2}{c}{} &Binary Verification &Type Identified &Critique Generation  \\ \midrule
         \multicolumn{2}{l}{Correct} & 78/100 (78\%) & 78/100 (78\%) & 78/100 (78\%) \\ \midrule
         \multicolumn{2}{l}{Inexecutable} & 70/100 (70\%) & 63/100 (63\%) & 8/100 (8\%) \\ \midrule
         \multicolumn{2}{l}{Non Goal Reaching} & 98/100 (98\%) & 12/100 (12\%) & 12/100 (12\%) \\ \midrule
         \multicolumn{2}{l}{Random}& 100/100 (100\%)& 94/100 (94\%)& 2/100 (2\%) \\ \midrule
        \multicolumn{1}{c}{\multirow{3}{*}{LLM}} & Correct & 53/55 (96.36\%) & 53/55 (96.36\%) & 53/55 (96.36\%) \\ \cmidrule{2-5}
         & Inexecutable & 25/40 (62.5\%) & 24/40 (60\%) & 0/40 (0\%) \\ \cmidrule{2-5}
         & Non Goal Reaching & 3/5 (60\%) & 2/5 (40\%) & 2/5 (40\%) \\ \midrule
         \bottomrule
         
    \end{tabular}
    \caption{This table presents the verification and critique accuracy of LLM-as-verifier across five different kinds of plans over 100 instances of Blocksworld.}
    \label{tab:blocksworld}
\end{table}

\begin{table}[]
    \centering
    \begin{tabular}{lllll}
    \toprule
         \multicolumn{2}{c}{} &Binary Verification &Type Identified &Critique Generation  \\ \midrule
         \multicolumn{2}{l}{Correct} & 3/100 (3\%) & 3/100 (3\%) & 3/100 (3\%) \\ \midrule
         \multicolumn{2}{l}{Inexecutable} & 100/100 (100\%) & 100/100 (100\%) & 24/100 (24\%) \\ \midrule
         \multicolumn{2}{l}{Non Goal Reaching} & 98/100 (98\%) & 12/100 (12\%) & 12/100 (12\%) \\ \midrule
         \multicolumn{2}{l}{Random}& 100/100 (100\%)& 100/100 (100\%)& 59/100 (100\%) \\ \midrule
        \multicolumn{1}{c}{\multirow{3}{*}{LLM}} & Correct & 0/3 (0\%) & 0/3 (0\%)& 0/3 (0\%) \\ \cmidrule{2-5}
         & Inexecutable & 89/89 (100\%) & 89/89 (100\%) & 12/89 (13.48\%) \\ \cmidrule{2-5}
         & Non Goal Reaching & 8/8 (100\%) & 0/8 (0\%) & 0/8 (0\%) \\ \midrule
         \bottomrule
         
    \end{tabular}
    \caption{This table presents the verification and critique accuracy of LLM-as-verifier across five different kinds of plans over 100 instances of Mystery Blocksworld.}
    \label{tab:mystery}
\end{table}
\subsubsection{Prompts - Blocksworld}
All of following examples are built on the same graph instance.
\mytcbinput{appendix/blocksworld/base_prompt.tex}{Baseline{,} Direct Prompt}{0}{prompt_bw:baseline}
\mytcbinput{appendix/blocksworld/example_response.tex}{Example LLM Response}{0}{example_bw:response}
\mytcbinput{appendix/blocksworld/llm_v_prompt.tex}{Prompt To Elicit Verification (Open Ended)}{0}{prompt_bw:verification}
\mytcbinput{appendix/blocksworld/llm_verification.tex}{Prompt To Elicit Verification (Format based)}{0}{prompt_bw:verification_format}
\mytcbinput{appendix/blocksworld/llm_v_cot.tex}{Prompt To Elicit Verification (Chain of thought based)}{0}{prompt_bw:verification_cot}
\mytcbinput{appendix/blocksworld/llm_v_swap.tex}{Prompt To Elicit Verification (Swapping Answer and Reason Order)}{0}{prompt_bw:verification_swap}

\subsubsection{Backprompts - Blocksworld}\label{appendix:mb-backprompts}
\mytcbinput{appendix/blocksworld/llm-self_example_backprompt.tex}{Backprompt Generated From Self-Critique}{0}{backprompt_bw:self}
\mytcbinput{appendix/blocksworld/first_example_backprompt.tex}{Backprompt Generated From Sound Verifier (First)}{0}{backprompt_bw:first}
\mytcbinput{appendix/blocksworld/full_example_backprompt.tex}{Backprompt Generated From Sound Verifier (Full)}{0}{backprompt_bw:full}

\subsubsection{Complete Example of Iterative Prompt Sequence - Blocksworld}
\mytcbinput{appendix/blocksworld/full_trace_example.tex}{Full Sequence of (LLM-Generated) Backprompts}{0}{example_bw:trace}

\subsubsection{LLM as Verifier - Blocksworld}
\mytcbinput{appendix/blocksworld/LLM_incorrect_meet_of_preconditions.tex}{Examples of LLM Output on the Verification Task}{0}{example_bw:normal_v_responses}

\subsubsection{Prompts - Mystery Blocksworld}
All of following examples are built on the same graph instance.
\mytcbinput{appendix/planning/base_prompt.tex}{Baseline{,} Direct Prompt}{0}{prompt_mb:baseline}
\mytcbinput{appendix/planning/example_response.tex}{Example LLM Response}{0}{example_mb:response}
\mytcbinput{appendix/planning/llm_v_prompt.tex}{Prompt To Elicit Verification (Open Ended)}{0}{prompt_mb:verification}
\mytcbinput{appendix/planning/llm_verification.tex}{Prompt To Elicit Verification (Format based)}{0}{prompt_mb:verification_format}
\mytcbinput{appendix/planning/llm_v_cot.tex}{Prompt To Elicit Verification (Chain of thought based)}{0}{prompt_mb:verification_cot}
\mytcbinput{appendix/planning/llm_v_swap.tex}{Prompt To Elicit Verification (Swapping Answer and Reason Order)}{0}{prompt_mb:verification_swap}
\subsubsection{Backprompts - Mystery Blocksworld}\label{appendix:mb-backprompts}
\mytcbinput{appendix/planning/llm-self_example_backprompt.tex}{Backprompt Generated From Self-Critique}{0}{backprompt_mb:self}
\mytcbinput{appendix/planning/first_example_backprompt.tex}{Backprompt Generated From Sound Verifier (First)}{0}{backprompt_mb:first}
\mytcbinput{appendix/planning/full_example_backprompt.tex}{Backprompt Generated From Sound Verifier (Full)}{0}{backprompt_mb:full}

\subsubsection{Complete Example of Iterative Prompt Sequence - Mystery Blocksworld}
\mytcbinput{appendix/planning/full_trace_example.tex}{Full Sequence of 15 (LLM-Generated) Backprompts}{0}{example_mb:trace}

\subsubsection{LLM as Verifier - Mystery Blocksworld}
\mytcbinput{appendix/planning/LLM_incorrect_meet_of_preconditions.tex}{Examples of LLM Output on the Verification Task}{0}{example_mb:normal_v_responses}

\end{document}